\documentclass[letterpaper,10pt,conference]{ieeeconf}
\IEEEoverridecommandlockouts

\usepackage[T1]{fontenc}
\usepackage{cite}
\usepackage{amsmath,amssymb,amsfonts}
\usepackage{algorithmic}
\usepackage{graphicx}
\usepackage{textcomp}
\usepackage{xcolor}
\usepackage{subcaption}
\def\BibTeX{{\rm B\kern-.05em{\sc i\kern-.025em b}\kern-.08em
    T\kern-.1667em\lower.7ex\hbox{E}\kern-.125emX}}

\usepackage{booktabs}
\usepackage[export]{adjustbox}
\let\labelindent\relax
\usepackage{enumitem}
\usepackage{wrapfig}
\usepackage{multirow}
\usepackage{tabularx}

\usepackage[noend,algo2e,ruled,vlined]{algorithm2e} 
\usepackage{algorithm}

\SetKwInOut{InOutput}{In Out}
\SetKwInOut{Input}{In}
\SetKwInOut{Output}{Out}

\newtheorem{definition}{Definition}[section]

\newcommand{\tool}{{\sc CaIdLogic\ }}

\usepackage[hidelinks]{hyperref}
\begin{document}

\title{Automatic Cause Determination in Road Scene Understanding Using Qualitative Reasoning and Four-Valued Logic
}
\author{Nassim Belmecheri$^{1}$ \and Arnaud Gotlieb$^{1}$ \and Nadjib Lazaar$^{2}$ \and Helge Spieker$^{1}$%
\thanks{$^{1}$Simula Research Laboratory, Oslo, Norway {\tt\small \{nassim,arnaud, helge\}@simula.no}}%
\thanks{$^{2}$LISN, Université Paris-Saclay, Saclay, France
        {\tt\small lazaar@lisn.fr}}%
\thanks{This work is funded by the European Commission through the AI4CCAM project (Trustworthy AI for Connected, Cooperative Automated Mobility) under grant agreement No 101076911, and by the Research Council of Norway through the AutoCSP project, grant number 324674.}%
}

\IEEEaftertitletext{%
\begin{minipage}{\textwidth}\footnotesize
Author manuscript of: N. Belmecheri, A. Gotlieb, N. Lazaar, and H. Spieker,
``Automatic Cause Determination in Road Scene Understanding Using Qualitative Reasoning and Four-Valued Logic,''
\textit{2025 IEEE Intelligent Vehicles Symposium (IV)}, pp. 1056--1061, 2025.
DOI: \href{https://doi.org/10.1109/IV64158.2025.11097599}{10.1109/IV64158.2025.11097599}.\endgraf
\smallskip
\textcopyright\ 2025 IEEE. Personal use of this material is permitted. Permission from IEEE must be obtained for all other uses, in any current or future media, including reprinting/republishing this material for advertising or promotional purposes, creating new collective works, for resale or redistribution to servers or lists, or reuse of any copyrighted component of this work in other works.
\end{minipage}\endgraf\vspace{1em}}
\maketitle
\begin{abstract}
Road scene understanding in automated driving (AD) aims to build a comprehensive analysis of video sequences taken on the road by embedded or fixed cameras (e.g., mounted on vertical road signals). One goal is to identify the relevant actors in the scene and another goal is to determine the causes that have triggered a specific action of the ego car (i.e., stop, slow down, turn left, etc.). In a complex urban environment, 
these causes can be multiple, confusing, possibly contradictory to other causes and not easily expressible using simplistic reasoning. Still, providing accurate {\it automatic cause determination} supports a) user acceptance by providing appropriate explanations to the car passengers and road users; b) increased road safety by providing detailed road scene understanding to traffic. In this paper, we propose using spatiotemporal reasoning and Belnap's four-valued logic to formulate complex causes of AD action in a road scene. We compute these causes by analysing a Qualitative eXplainable Graph (QXG), which is an abstract representation of the road scene capturing spatiotemporal relations between road entities. Starting from a QXG, our approach called \tool, is targeted to determine complex causes of a selected AD action occurring in a specific frame of a road scene. The usefulness of \tool is demonstrated on several scenes extracted from the well-known NuScene dataset.     
\end{abstract}

\section{Introduction}
\begin{figure*}[h!]
    \centering
    \begin{subfigure}{0.32\textwidth}
        \centering
        \includegraphics[width=0.88\linewidth]{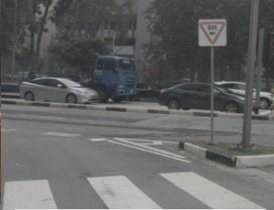}
        \caption{Ego car stops because it is approaching a crossing line and there is a give way sign.\\ \tool~: {\it c1.sort=infra\_stop, c1.fvl=T}}
        \label{fig:im1}
    \end{subfigure}
    \hfill
    \begin{subfigure}{0.32\textwidth}
        \centering
        \includegraphics[width=0.99\linewidth]{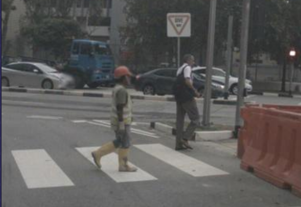}
        \caption{Ego car stops because there is a ped crossing on the upfront zebra.\\
        \tool~: {\it c2.sort=inroad\_ped, c2.fvl=T}}
        \label{fig:im2}
    \end{subfigure}
    \hfill
    \begin{subfigure}{0.32\textwidth}
        \centering
        \includegraphics[width=0.85\linewidth]{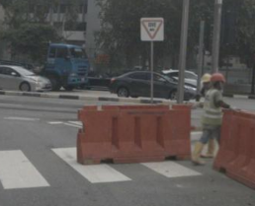}
        \caption{Ego car stops because there is a static object.\\
        \tool~: {\it c3.sort=inroad\_stat\_obj, c3.fvl=T}\\
        \tool~: {\bf c2.fvl=B}}
        \label{fig:im3}
    \end{subfigure}
    \caption{Cause determination is complex in urban traffic scenarios.}
    \label{fig:complex_causes}
\end{figure*}

In automated driving (AD), {\it road scene understanding} is paramount to explain a situation where an ego car has taken a particular action. By identifying the relevant actors of a scene such as its vehicles, road users (i.e. pedestrians, cyclists, and so on) and relevant objects, and determining the causes that have triggered a specific action of the ego car, road scene understanding supports car passengers, traffic managers, and other road users to better accept AD systems on the road and to make them safer \cite{xue_survey_2018}. 
However, determining the cause of an ego car's action in a road scene is a complex task that involves multiple factors. The ego car's decisions are influenced by dynamic elements such as traffic flow, pedestrian movement, road conditions, and surrounding vehicles' behaviors. Each of these elements can change unpredictably and the integration of various sensors, including cameras, LiDAR, and radar, generates vast amounts of data that must be accurately processed and interpreted in real-time. For example, consider the scenes given in Fig.\ref{fig:complex_causes}. In the three cases, the ego car stops but automatically determining the exact cause of that action is difficult because the causes are multiple and ambiguous.  
Understanding the context of the road scene, such as interpreting traffic signals, road signs, and the intentions of other road users, adds another layer of complexity. Determining the precise cause of an ego car's action is an important but also challenging and multifaceted problem.

Existing automatic cause determination of ego car actions can be categorized into vision-based and knowledge-based methods \cite{Kim2018}. Vision-based methodologies use ML models trained on human-annotated datasets, often derived from vision-language models like Visual Query Answers (VQAs), and generate explanations through text produced by generative AI approaches \cite{inoue_nuscenes-mqa_2023}. These methods are adaptable to new situations but are inherently non-deterministic, posing challenges for operational deployment. Knowledge-based methods, on the other hand, rely on causality and causal models, using knowledge and causal graphs to generate explanations, and can deterministically compute the cause of actions without relying on potentially biased datasets \cite{wormann_knowledge_2023}. A recently introduced method, called sceneXplain, has defined Qualitative eXplainable Graphs (QXGs) to exploit qualitative reasoning for scene understanding \cite{belmecheri2023acquiring}. However, this method cannot compute the causes of ego car actions, as it is restricted to object identification and ego car action determination. 

In this paper, we propose a novel knowledge-based method leveraging spatiotemporal reasoning combined with Belnap's four-valued logic (FVL)  to formulate complex causes of autonomous driving (AD) actions within a road scene. These causes are computed by analyzing a QXG, which serves as an abstract representation of the road scene, capturing the spatiotemporal relationships between road entities. Our approach, named \tool, is designed to identify complex causes of a selected AD action occurring within a specific frame of a road scene. The effectiveness of \tool is demonstrated through several scenarios from the well-known NuScenes dataset \cite{nuscenes2019}.  

In Fig.\ref{fig:complex_causes}, the three captions not only give the causes of a stopping action of the ego car in plain text, but also show the computed results of \tool. In the first and second subfigures, true causes are identified (i.e., {\it infra\_stop} and {\it inroad\_ped}), while another true cause is identified in the third subfigure ({\it inroad\_static\_obj}). However, in this third subfigure, \tool revises the truth value of the second cause ({inroad\_ped}) as the pedestrian is not the main cause for the vehicle to stop. 
The rest of the paper is organized as follows: 
Sec.\ref{sec:back} presents the necessary background required to understand the paper. Sec.\ref{sec:cause} describes the process of automatic cause identification using FVL. Sec.\ref{sec:implem} introduces our implementation \tool, and discusses the initial results obtained by applying \tool to a subset of scenes extracted from the NuScenes dataset. Finally, Sec.\ref{sec:concl} concludes the paper and outlines future perspectives.

\section{Background}
\label{sec:back}

\noindent {\bf Road Scene Representation.}
A road scene is a sequence of $n$ frames $\mathcal{V}=\langle f_1,\ldots, f_n \rangle$ representing the bird-eye-view of an AD ego car. In each frame, a set of entities $\mathcal{E}$ can be automatically detected and tracked \cite{JanaiGBG20,scene_graph_survey}. \\

\noindent {\bf Qualitative Calculus.}
A \emph{qualitative calculus} is a method that computes the qualitative relationships among physical attributes of entities, such as relative positions, velocity, etc ~\cite{dylla_survey_2015}. These qualitative relations do not rely on exact quantitative evaluation. Qualitative calculus is based on different algebras useful to capture spatial and/or temporal relationships between entities~\cite{allen_maintaining_1983,renz_qualitative_2007}. \\

\noindent {\bf Qualitative Algebras.}
Regarding spatial relationships, multiple complementary qualitative algebras can be used. We just explain two of them which are well-suited in the context of AD:
\begin{itemize}
\item    Qualitative Distance Calculus (QDC) \cite{renz_qualitative_2007}: Dedicated to estimate the distances between entities, this algebra determines if they are "close", "very close", "far" or "very far" without establishing any precise measurements. In AD, this algebra is particularly well-suited to express distance-based relationships;
\item    Rectangle Algebra (RA) \cite{renz_qualitative_2007}: In this algebra, each entity is represented by a rectangle corresponding by its detected bounding box. Relations between entities are then constructed by using the thirteen relations of the Allen's interval algebra~\cite{allen_maintaining_1983}, which are applied to both the x- and y-axes. These relations (i.e., \textit{before, meets, overlaps, during, starts, finishes, equals, finished by, started by, overlapped by, met by, after, and contains}) allow us to compute relative entity positioning.
\end{itemize}

\begin{figure*}
    \centering
    \begin{tabular}{|c|c|}\hline
           \includegraphics[scale=0.50]{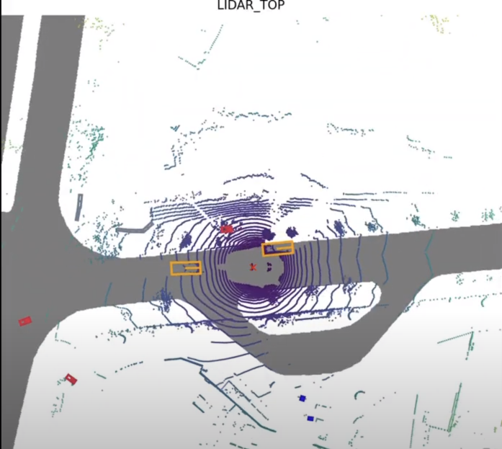}
  &     \includegraphics[scale=0.80]{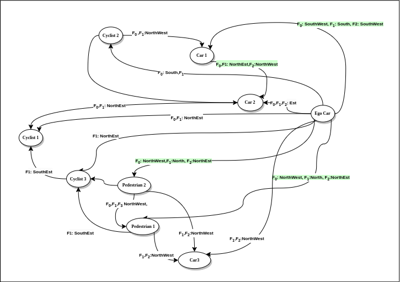}\\
  \hline
    \end{tabular}
    \caption{A complex road scene and its computed QXG}    \label{fig:qxg}
\end{figure*}

\noindent {\bf Qualitative eXplainable Graph (QXG).}
In \cite{belmecheri2023acquiring}, we introduced a novel representation of a road scene, called the Qualitative eXplainable Graph (QXG). Formally, for a given scene $\mathcal{V}$, the QXG is a graph $\mathcal{G}=(\mathcal{E}, \mathcal{R})$, where $\mathcal{E}$ is a finite set of entities found in $\mathcal{V}$, and $\mathcal{R}$ is a set of edges labeled by the qualitative relations between the corresponding pairs of entities. 
Fig.\ref{fig:qxg} shows the QXG constructed from the AD road frame. The scene depicts an ego car with multiple entities which are liaised with the ego through different spatiotemporal relation as depicted in the corresponding QXG. We refer the interested reader to \cite{belmecheri2025trustworthy} for more details. 

\begin{figure}
  \centering
    \includegraphics[width=0.4\linewidth]{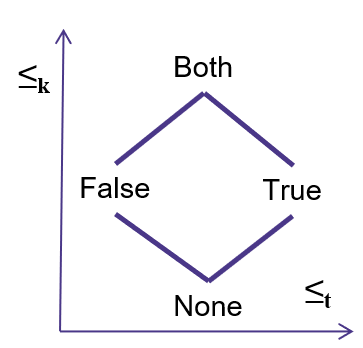}
    \caption{Bilattice Structure}
\label{fig:bilat}
\end{figure}

\begin{figure}
    \centering
    \begin{tabular}{|c|c|}\hline
        \includegraphics[scale=0.07]{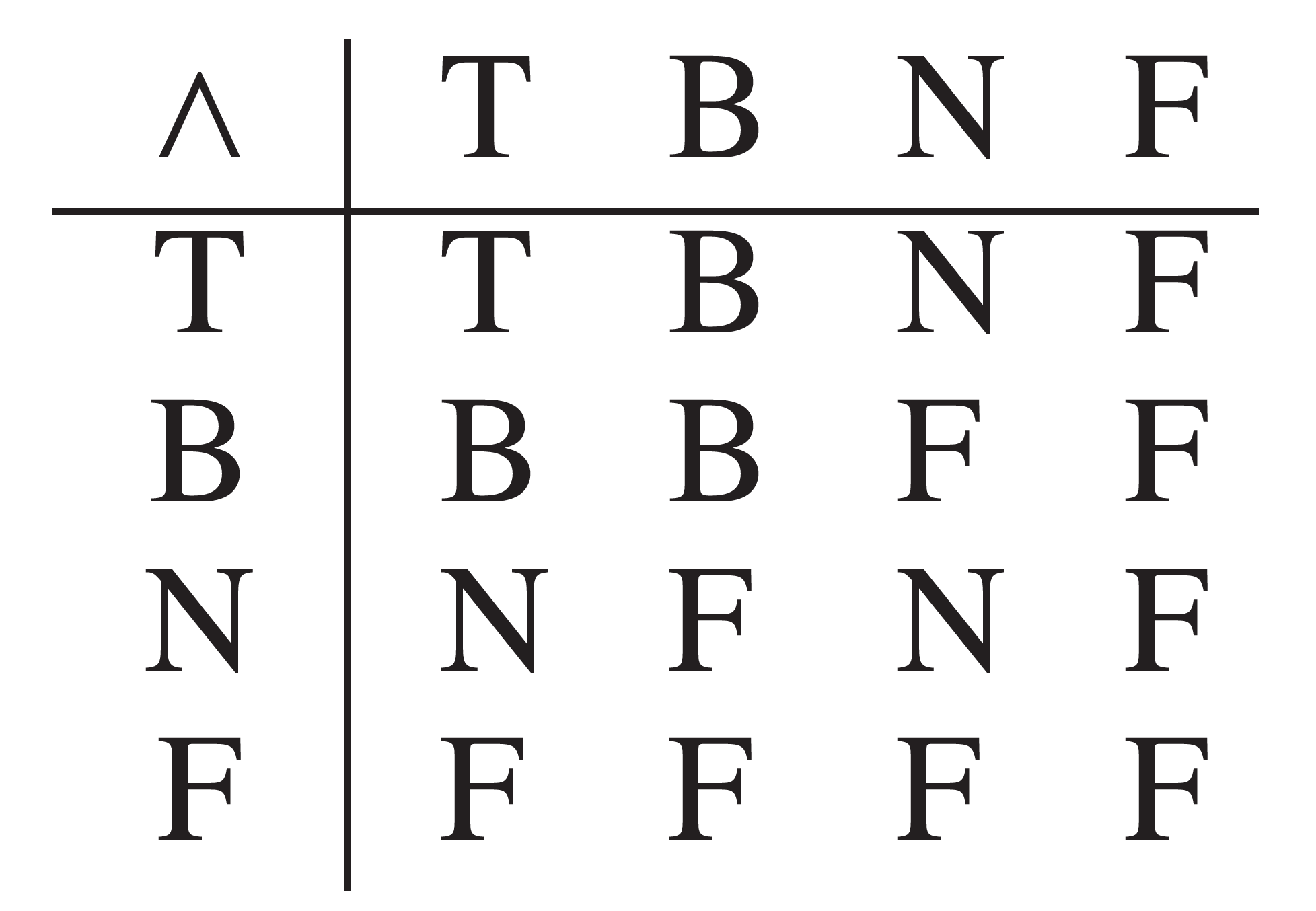}
  &     \includegraphics[scale=0.07]{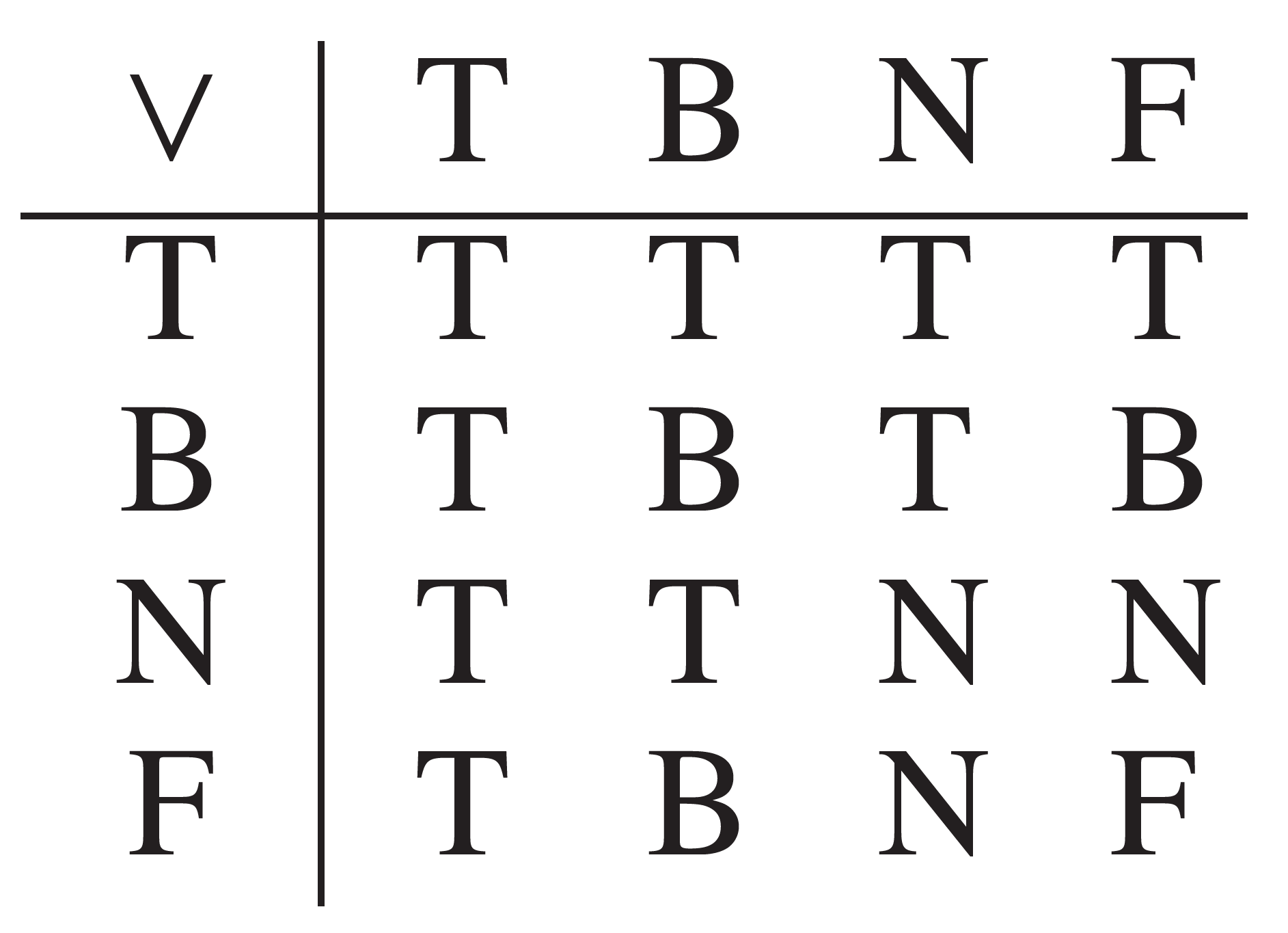}\\
        \includegraphics[scale=0.7]{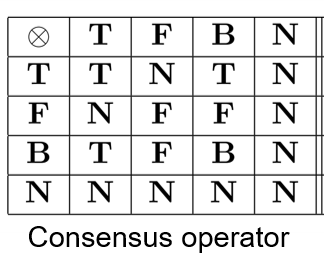}
  &     \includegraphics[scale=0.7]{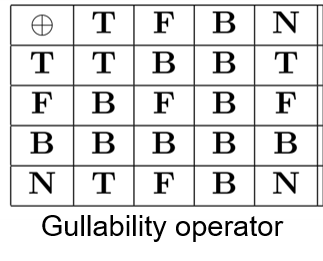}\\
    \hline
    \end{tabular}
    \caption{Operators in Belnaps's Four-Valued Logic}    \label{fig:op}
\end{figure}

\noindent{\bf Belnap's Four-Valued Logic (FVL).}
Belnap's {\it Four-valued logic}\footnote{Note that there are other four-valued logics, such as the Quantum four-value logic, which appear as being less relevant for our AD context.} extends classical two-valued first-order logic by introducing four truth values: True ($T$), False ($F$), Both ($B$), and Neither ($N$) \cite{belnap77}. This framework is beneficial in dealing with inconsistent or incomplete information. The four values represent:
\begin{enumerate}
\item[$T$:] The cause is known and true;
\item[$F$:] The cause is known and false;
\item[$B$:] The cause is known and both true and false, capturing contradictions
\item[$N$:] The statement is unknown, neither true nor false, capturing lack of information.
\end{enumerate}
FVL is equipped with a bi-lattice structure over $\mathcal{C}$,
a set of FVL-clauses, where $T$ and $F$ are incomparable, while $B$ is the supremum (least upper bound) and $N$ is the infimum (greatest lower bound) \cite{lallouet_2005}, as shown in Fig.\ref{fig:bilat}. 
This logic is applied in areas where information may be incomplete or conflicting \cite{belnap77}.
In FVL, clauses can be constructed as usual in first-order logical clauses using variables, terms and operators with the difference that they can take either of the four values ${T,F,B, N}$. By default, any FVL-clause $c \in \mathcal{C}$ can be combined with other FVL-clauses using operators. The primary logical operators include $\wedge$ and $\vee$ as shown in Fig.\ref{fig:op}.
In addition to these standard operators, FVL introduces two other binary operators, namely gullibility and consensus. 
Gullibility, noted $\oplus$ reflects a tendency to accept information as true unless explicitly false, while Consensus, noted $\otimes$ represents a cautious approach, accepting information as true only if it is consistently true. The full definition tables for these operators are given in Fig.\ref{fig:op}. These operators are particularly useful in decision-making processes where information may be ambiguous. Gullibility favors the acceptance of uncertain information, while consensus favors rejection unless certainty is achieved. This makes FVL a powerful tool for dealing with contradictory causes, incomplete information related to causes identification, etc.

\section{The Cause Identification Problem}
\label{sec:cause}
In this paper, we address the following problem which aims at automatically detecting the cause of an ego car action: 
\begin{definition}[Cause Identification Problem]
Given a video scene $\mathcal{V}$, an ego car action $a$ occurring at frame $f_k\in \mathcal{V}$, an horizon of time $h$ and a set of possible causes $\mathcal{C}$, the {\it Cause Identification Problem} consists in computing a conjunction of logical FVL-labeled clauses $c_1 \wedge \ldots \wedge c_n$ which identify the causes from $\mathcal{C}$ and the involved entity of $a$, found in the $h$ last successive frames $f_{k-h}, f_{k-h+1}, \ldots,f_{k-1}$.  
\end{definition}

Several remarks underline the choices made in this definition:
\begin{itemize}
    \item We are interested in causes of ego car actions: It explains why the potential causes can be found in a pre-determined set of logical clauses $\mathcal{C}$~;
    \item A cause is sought by looking backwards from a given frame $f_k$ where the action of the ego car takes place. For that, a horizon of time $h$ is preselected that determines how much frames will be examined $f_{k-h}, f_{k-h+1},\ldots,f_{k-1}$. Note that, behind the choice of $h$, there is a strong hypothesis which can be framed as follows: "the cause of $a$ can be found in the $h$ frames preceding $f_k$". This is debatable (the actual cause could be found at more than $h$ frames) even though some consensus from the state-of-practice tends to say that any action of an ego car depends on the $5$ preceding frames. In our experiments, we explore a value of $h \in 3..10$~;
    \item A Cause $C$ is defined as a {\it "conjunction of logical FVL-clauses"} $c_1 \wedge \ldots \wedge c_n$ where each $c_i$ is a structured data containing a $c.sort$ (which sort of cause is concerned), an $c.fvl$ value among $B, T, F, N$, the first frame where the cause has been identified $c.frame$ and the entity that is identified as cause responsible. This data choice allows us to consider incomplete or partial knowledge while simultaneously trying to identify precisely the cause of the EGO car action. It also accounts for the difficulty of affirming that something is the true cause of an event. 
    \item As argued in the introduction, considering that causes have an FVL value is an important, but also tricky aspect. It raises the question of combining the multiple causes found in different frames. We have decided to collect and combine causes of the same sort using the FVL. A discussion is raised in Sec.\ref{sec:implem} about this choice. 
\end{itemize}

\subsection{Causes of EGO Car Actions}
Automated vehicles take actions based on a complex array of sensors and algorithms that analyse their environment. Several key triggers (called "sort of cause" in our paper) can prompt these vehicles to take action. One primary cause is the detection of a pedestrian on the road, where the vehicle identifies a pedestrian on the road within a specific frame of reference. This cause is represented as the following FVL-clause {\it c.sort = inroad\_ped} in our framework, which identifies the first frame ($c.frame = f_i$) where the tracked entity (a unique identifier for the pedestrian $c.entity = id\_ped$) has been detected. In our framework, this scenario necessitates immediate braking to avoid a collision, which is called {\tt going\_straighthand\_driving\_not\_moving}. Note that we currently only deal with this action in our framework but other founding causes of other ego car actions is also paramount. Similarly, the presence of a stopped vehicle on the road, called {\it c.sort = inroad\_front\_stopped\_vehicle}, where an identified car ($c.entity = id\_car$) is detected as stopped directly ahead, requires the automated vehicle to stop to prevent an accident. Another critical trigger is the identification of a static object on the road, called {\it c.sort = inroad\_static\_obj}, which involves recognizing any stationary object on the road ($c.entity = id\_obj$), such as debris or construction barriers, prompting the vehicle to navigate around it safely. The detection of a vehicle parked roadside and starting to engage in the main road requires the EGO car to stop or to change line ({\it c.sort = parked\_roadside\_vehicle}) to maintain a safe distance. Additionally, encountering a STOP road sign ({\it c.sort = infra\_stop}), such as a stop sign or traffic light as part of the road infrastructure, mandates the vehicle to halt and comply with traffic regulations.

Beyond these primary causes, automated vehicles must also respond to dynamic and complex scenarios. For instance, the detection of a moving vehicle ({\it c.sort = inroad\_moving\_vehicle}), where another vehicle is moving erratically or changing lanes, requires adaptive speed adjustments and potential lane changes. The presence of animals on the road ({\it c.sort = inroad\_animal}), such as a deer crossing the road, triggers emergency braking to avoid harm. Similarly, identifying a cyclist or any vulnerable road user ({\it c.sort = inroad\_cyclist}) necessitates careful navigation to ensure the cyclist's safety. Environmental conditions also play a significant role. For example, difficult weather conditions ({\it c.sort = adverse\_weather\_conditions}), such as heavy rain or fog, require the vehicle to adjust its speed and sensor sensitivity to maintain visibility and traction. The detection of issues related to the concrete used in the road ({\it c.sort = road\_surface\_issues}), like potholes, prompts the vehicle to adjust its path to avoid damage or loss of control. Additionally, recognizing any road signs ({\it c.sort = road\_signs}), such as speed limits or yield signs, ensures the vehicle adheres to legal and safe driving practices. Automated vehicles must also account for high traffic density ({\it c.sort =  traffic\_density}), where high traffic volumes necessitate adaptive cruise control and lane-keeping to navigate congested roads efficiently. The detection of emergency vehicles ({\it c.sort = emergency\_vehicles}), such as ambulances or fire trucks, requires the vehicle to yield and provide a clear path. Furthermore, recognizing road construction area ({\it c.sort = road\_construction\_zones}) prompts the vehicle to slow down and follow temporary traffic patterns.
By integrating these diverse causes, automated vehicles can explain their actions and more generally their navigation in complex environments with a high degree of safety and reliability.

\subsection{FVL-based Cause Calculation}
\begin{algorithm2e*}
\caption{FVL-based Cause Calculation}\label{alg:cc}	
 	\Input{QXG $(\mathcal{PS},\mathcal{FS},\mathcal{RS})$; {\it \ \ // $\mathcal{PS}$: Pairs of entities, $\mathcal{FS}$: Succssives frames, $\mathcal{RS}$: Qualitative relations for each pair}\\ 
           $f_k$; {\it \ \ // The frame when the EGO car action took place ($f_k \in \mathcal{FS}$)}\\
           $h$; {\it \ \ // Time horizon ($h \in 1..k$)}\ \ \ \
           $PosCauses$; {\it \ \ // A set of possible individual causes}
         }
	\Output{$nil$ or $\mathcal{C} = c_1 \wedge \ldots \wedge c_n$; {\it \ \ // Conjunction of FVL-clauses Interpreted as the cause of the action}  }

\BlankLine
$\mathcal{C} \gets nil$\;
\ForEach{$(e, ego) \in \mathcal{PS}$ \it \ \ // Only the entities interacting with the ego car are selected}{
\lForEach{$c_i \in PosCauses$}{$c_i.sort \gets nil, c_i.fvl \gets N$  \it \ \ // Initialization phase over all the possible clauses}
\ForEach{$f_j \in \mathcal{FS}: k-h \leq j \leq k-1$ and $c_i \in PosCauses$ \it \ \ // each frame is examined from $k$, up to $h$}
{
    $c \gets PatternCauseIdentify(e, f_j, \mathcal{RS}, PosCauses)$;  {\it  \ \ // In $f_j$, a cause related to entitity $e$ is identified }
    \If{$c \neq nil$ and $c.sort = c_i.sort$}{ %
    $c_i.fvl \gets c_i.fvl \oplus c.fvl$; \ \    $c_i.frame \gets add(f_j, c_i.frame)$; \ \ $c_i.entity \gets e$\;
    {\it \ \ // Causes of the same sort, related to the same entity} are combined using gulability}
    $\mathcal{C} \gets \mathcal{C} \wedge c_i$\;
}
}
\Return{$\mathcal{C}$}\;
\end{algorithm2e*}
This section presents a first algorithm aiming at calculating the cause of an ego car action, and thus solving the Cause Identification problem. The algorithm is quite simple, using successive and imbricated iterations on the objects which interact with the ego car (first loop), initialisation of the FVL-based clauses (second iteration), and backward exploration of the frames until the horizon is fully explored (third iteration). When a cause pattern is identified (function $PatternCauseIdentify$), it is then combined with the causes of the same sort previously identified using the gullability operator. Choosing this operator is crucial as it tends to estimate precisely the possible causes of a given action. In the absence of ground truth, this choice appears as the most reasonable. Alternatively, choosing the consensus operator would have led to often overestimate the uncertainty in cause identification in our opinion.  
The crucial function $PatternCauseIdentify$ identifies known patterns in $\mathcal{RS}$ for specific causes. This function can either be implemented using logical pattern recognition (as done in our framework) or be learned from existing labeled datasets. Interestingly note that using FVL allows us to take advantage of not only both positive (true cause in this frame) and negative (false cause) detected cause patterns, but also uncertain patterns (both true and false cause in this frame). This can be extremely advantageous to represent uncertain knowledge such occluded objects or road users in a scene.

\section{Implementation, Initial Results and Discussion}
\label{sec:implem}

We implemented \tool as a tool that calculates the causes of ego car actions using logic programming (i.e., SICStus Prolog) and the Belnap's FVL. By processing a file containing the QXG representation of a given scene, \tool determines the appropriate action for the ego car. We defined predicates for each potential cause of action, such as {\tt inroad\_pedestrian(entity, frame)}, {\tt inroad\_front\_stopped\_vehicle(CAR, frame)}, and {\tt inroad\_static\_object(OBJ, frame)}. Each predicate is associated with a truth value from Belnap's logic (as shown in Alg.\ref{alg:cc}), allowing the system to capture the certainty or ambiguity of the information. The function {\it PatternCauseIdentify}, used in Alg.\ref{alg:cc}, is implemented using Prolog's logical inference capabilities based on usual pattern matching. To handle the gullibility and other operators of the FVL, we defined additional predicates able to handle any kind of FVL-formula. 
Even though \tool currently handles only a single car action, namely {\tt going\_straighthand\_driving\_not\_moving}, it is complete regarding the possible causes that have driven the car to take this action.
Our current Prolog implementation is composed of 450LOC. All our tests were performed on a standard PC with an Intel Core i7 processor running at 2.8GHz with 64GB RAM.

\subsection{Preliminary Results} 
To evaluate \tool, we selected $10$ scenes at random from the Nuscene dataset \cite{nuscenes2019}. Nuscene is an open-access dataset of $850$ annotated AD scenes. It is commonly used for benchmarking various AD-dedicated algorithms and it has been extended with several types of annotation \cite{inoue_nuscenes-mqa_2023}. As there is no annotation using FVL we had to identify the causes of actions manually to verify the result computed by \tool.
\begin{table*}
\centering\begin{adjustbox}{max width=\textwidth}
\begin{tabular}{ |p{0.8cm}||p{5.5cm}|p{10.5cm}|  }
 \hline
 {\bf Scene} & {\bf Manually identified cause from the QXG} & {\bf Cause computed by \tool} \\
 \hline
    26  & A traffic cone is very close to the ego car frame 5 to frame 9  & c1.sort = inroad\_movable, c1.fvl = B, c1.entity=, trafficcone, c1.frame = [9,8,7,6,5]\\
    \hline
    63  & No cause is identifiable in the scene & c1.sort = nil, c1.fvl = N\\
    \hline
    72  &  No cause is identifiable in the scene & c1.sort = nil, c1.fvl = N\ \\
    \hline
    174 &  A vehicule (car) is stopped in front of the ego car at frame 9 & c1.sort = inroad\_front\_stopped\_vehicle, c1.fvl = T, c1.entity = car, c1.frame = [9] \\   
    \hline
    177         &  A vehicule (truck) is stopped in front of the ego car at frames 15 and 16    & c1.sort = inroad\_front\_stopped\_vehicle, c1.fvl = T, c1.entity = truck, c1.frame = [16,15]\\
    \hline
    198        &  No cause is identifiable in the scene   & c1.sort = nil, c1.fvl = N\\
    \hline
    201       &  No cause is identifiable in the scene    & c1.sort = nil, c1.fvl = N \\
    \hline
    206         &  No cause is identifiable in the scene  & c1.sort = nil, c1.fvl = N\\
    \hline
    263         &  A vehicule (car) is stopped in front of the ego car at frame 0 and another vehicule (truck) s stopped n front of the ego car at frame 6  &  c1.sort = inroad\_front\_stopped\_vehicle, c1.fvl = T, c1.entity = car, c1.frame = [0] 
    c2.sort = inroad\_front\_stopped\_vehicle, c2.fvl = T, c2.entity = truck, c2.frame = [6] \\
    \hline
    268         &  A barrier is very close to the go car from frame 1 to 9 and there are 2 vehicules (car 1 and car 2) stopped in front of the ego car from frame 0 to frame 9 &  c1.sort = inroad\_movable, c1.fvl = B, c1.entity = barrier, c1.frame = [9,8,7,6,5,4,3,2,1]  c2.sort = inroad\_front\_stopped\_vehicle, c2.fvl = T, c2.entity = car(1), c2.frame = [9,8,7,6,5,4,3,2,1,0] c3.sort = inroad\_front\_stopped\_vehicle, c3.fvl = T, c3.entity = car(2), c3.frame = [9,8,7,6,5,4,3,2,1,0]\\
 \hline
\end{tabular}
\end{adjustbox}
\caption{Cause Identification using \tool. Horizon=10}\label{tab:res}
\end{table*}

The preliminary results presented in Tab.\ref{tab:res} obtained when applying \tool show that:
\begin{itemize}
\item As suspected, causes in AD scene are sometimes ambiguous as showed by the results for scenes 26 and 268 which both have 'B' as FVL-value;
\item Causes can be multiple as shown in scenes 263 and 268;
\item In many cases, there is no identifiable causes that can explain the action of the ego car, as in scenes 63, 72, 198, 201 and 206;
\end{itemize}
Note that any of these cause determination requests took less than one millisecond of CPU time to solve, so we have not reported on the time as it is negligible in these examples. As an explanation, the QXG representation is a frugal spatiotemporal representation of the video scene, which explains why it is so fast compared to searching the video itself.
\subsection{Discussion}
The preliminary results shown in this paper are insufficient to ensure complete coverage of the possible situations. In addition, the current limitations of \tool, especially regarding the pattern-matching capabilities of possible causes, do not allow us to draw strong conclusions when arguing the feasibility of automatic cause identification for very complex situations. However, using FVL to account for uncertainty and ambiguity in cause identification, as well as multiple causes is verified. 

\section{Related Work}
\label{sec:related}
Existing automatic cause determination of ego car actions can be categorized into two fields: vision-based or knowledge-based methods. 
\begin{itemize}
\item Vision-based methodologies leverage machine learning (ML) models trained on datasets annotated by humans \cite{kim_textual_2018}. These annotations are often derived from vision-language models, specifically labeled as Visual Query Answers (VQAs), which facilitate the acquisition of precise and accurate responses \cite{inoue_nuscenes-mqa_2023}. Subsequently, an explanation generator can articulate a scene by producing text generated from the computation of per-word softmax probabilities. These approaches do not necessitate a predefined knowledge base and exhibit adaptability to novel situations. However, their inherent non-deterministic nature poses challenges for deployment in operational environments.
\item In contrast to vision-based approaches, knowledge-based methods utilize knowledge and causal graphs to generate informed explanations \cite{wormann_knowledge_2023}. These methods fundamentally rely on causality and causal models to examine various aspects of autonomous driving (AD), including modality \cite{avetisyan_investigating_2022} and post-mortem complaint data \cite{ayoub_cause-and-effect_2022}. For example, in \cite{maier_causal_2023}, the authors manually develop structural causal models based on clustered variables, specifically designed to facilitate scenario-based testing of AD systems. As part of theses knowledge-based approaches, our methodology employs Qualitative eXplainable Graphs (QXGs), i.e., they utilize qualitative reasoning across multiple algebras to identify relevant objects within an AD scene \cite{belmecheri2025trustworthy} and to determine the actions of the ego vehicle \cite{belmecheri_roi_24}. The method has the advantage to deterministically computing the cause of the action, without relying on existing datasets which may be biased. It also allows us to reason with incomplete or uncertain knowledge.
\end{itemize}
\section{Conclusions and Perspectives}
\label{sec:concl}
By integrating Belnap's FVL into scene understanding, our approach makes a valuable step to effectively manage the uncertainties inherent in cause determination in AD systems. The result is an automated cause identification system that enhances the reliability and self-explanatory framework of the ego car's decision-making system. Our preliminary tests with \tool show that combining advanced FVL with Prolog's inference capabilities allows us to address the complexity of real-world cause determination in AD scenes. To the best of our knowledge, it is the first time Belnap's FVL is used in cause determination of ego car actions. Further work includes the implementation of cause determination for other ego car actions, including "slow down", as well as the extension of \tool with learned cause patterns. 
\section*{Acknowledgment}
Funded by the European Union through the AI4CCAM project (Trustworthy AI for Connected, Cooperative Automated Mobility) under grant agreement No 101076911. Views and opinions expressed are however those of the author(s) only and do not necessarily reflect those of the European Union or CINEA. Neither the European Union nor the granting authority can be held responsible for them.

\bibliographystyle{IEEEtran}
\bibliography{sceneunderstanding}

\end{document}